\documentclass[twocolumn,english]{article}
\usepackage[T1]{fontenc}
\usepackage[latin9]{inputenc}
\usepackage{color}
\usepackage{array}
\usepackage{amsmath}
\usepackage{graphicx}

\makeatletter

\providecommand{\tabularnewline}{\\}

\usepackage{babel}
\usepackage{caption}
\usepackage{iccv}
\usepackage{times}
\usepackage{epsfig}
\usepackage{cite}

\iccvfinalcopy 

\ificcvfinal\fi

\def\myDisplaySkip{0pt}
\abovedisplayskip=\myDisplaySkip
\belowdisplayskip=\myDisplaySkip
\abovedisplayshortskip=\myDisplaySkip
\belowdisplayshortskip=\myDisplaySkip

\makeatother

\begin{document}

\title{Material Recognition from Local Appearance in Global Context}

\author{Gabriel Schwartz\qquad{}\qquad{}Ko Nishino\\
Department of Computer Science, Drexel University\\
\texttt{\{gbs25,kon\}@drexel.edu}}
\maketitle
\begin{abstract}
Recognition of materials has proven to be a challenging problem due to the wide variation in appearance within and between categories. Global image context, such as where the material is or what object it makes up, can be crucial to recognizing the material. Existing methods, however, operate on an implicit fusion of materials and context by using large receptive fields as input (i.e., large image patches). Many recent material recognition methods treat materials as yet another set of labels like objects. Materials are, however, fundamentally different from objects as they have no inherent shape or defined spatial extent. Approaches that ignore this can only take advantage of limited implicit context as it appears during training. We instead show that recognizing materials purely from their local appearance and integrating separately recognized global contextual cues including objects and places leads to superior dense, per-pixel, material recognition. We achieve this by training a fully-convolutional material recognition network end-to-end with only material category supervision. We integrate object and place estimates to this network from independent CNNs. This approach avoids the necessity of preparing an impractically-large amount of training data to cover the product space of materials, objects, and scenes, while fully leveraging contextual cues for dense material recognition. Furthermore, we perform a detailed analysis of the effects of context granularity, spatial resolution, and the network level at which we introduce context. On a recently introduced comprehensive and diverse material database \cite{Schwartz2016}, we confirm that our method achieves state-of-the-art accuracy with significantly less training data compared to past methods.
\end{abstract}

\section{Introduction}
\begin{figure}
\begin{centering}
\includegraphics[width=0.98\columnwidth]{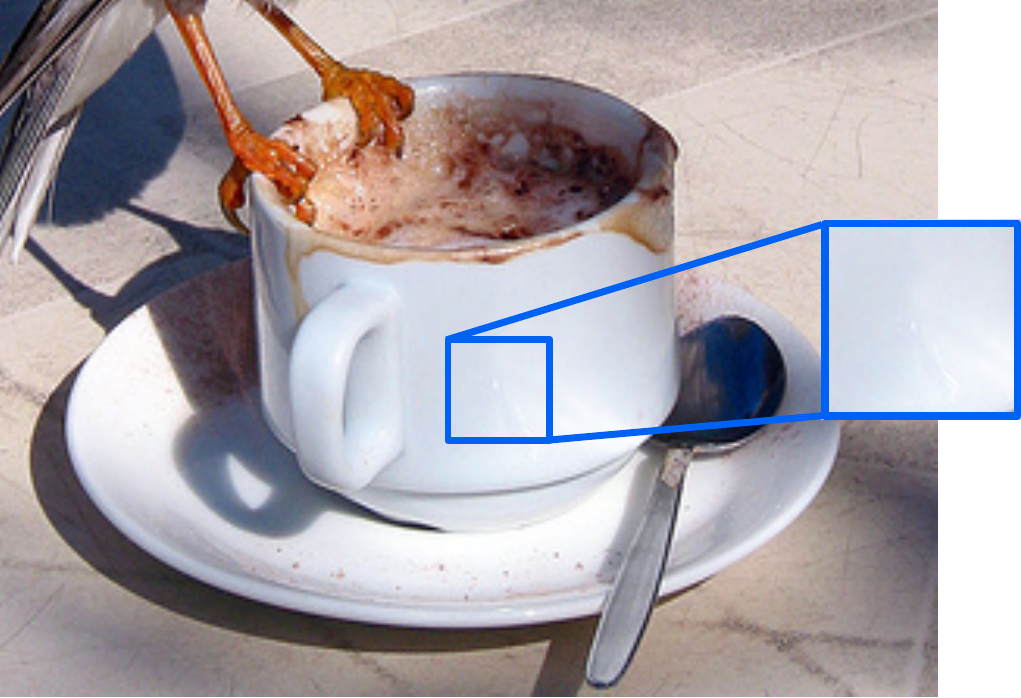}
\par\end{centering}
\caption{\label{fig:material_patch_zoom}Material recognition requires accurate
sources of both local and global information. Even if a material is
not uniquely determined by local appearance, local cues can constrain
the choice of possible materials. Here, the local appearance suggests
a smooth white material such as ceramic or plastic, and the global
appearance is that of a mug. Rather than mix these two sources of
information implicitly by just using large patches as input, we separate and then combine local materials and global object and scene context to improve material recognition.}
\end{figure}

Material recognition is an inherently challenging problem, primarily due to the large variation in appearance between different instances of a given material and between different materials. There has been significant recent progress in terms of accuracy on benchmark datasets. Most methods proposed to
achieve such results, however, essentially treat material recognition as object recognition with different categories. They often use large image patches that cover parts or whole objects and scenes, as one would when performing object recognition, which inevitably mix visual cues of materials and other image context, mainly of objects.

Material recognition is fundamentally different from object recognition. Adelson~\cite{Adelson2001}~alludes to this difference in his discussion of ``things'' vs.~``stuff''. While materials underlie both ``things'' and ``stuff,'' the key difference between them highlights the critical difference between objects and materials. Similar to ``stuff'', and unlike ``things'' (objects), materials may not necessarily be recognized by having a particular shape. A cup, an object with a typical cylindrical shape, is often made of ceramic, a material. The fact that an object is a cup can be used as a cue to recognize the material as ceramic, and likewise the presence of ceramic may suggest that an object might be a cup. ``Cup'' and ``ceramic'' are not, however, interchangeable. Not all ceramic ``things'' are cups and relying on shape cues to recognize ceramic, which past methods inevitably do, is a fundamentally limited approach. 

To avoid this uncontrolled dependence on context, materials need to be recognized locally (i.e., without seeing the object it makes up or the scene in which it lies), before other ``global'' context including objects and scenes recognized separately can help eliminate remaining uncertainty arising from strictly local appearance cues. Recognition of materials from local appearance cues becomes even more essential when recognizing ``stuff'' such as towels, water, and bushes that do not have canonical shapes.

Mixing context and material categories during material recognition implicitly relies on the underlying recognition framework to disentangle these concepts. Bell~\etal\cite{Bell2015} come to the conclusion that, ``Training on a dataset which includes the surrounding context is crucial for real-world material classification,'' but when simply given large patches as input and materials as output, we may only speculate as to the actual importance of context. In fact, Fig. 10 of \cite{Bell2015} clearly shows that materials are recognized by identifying the actual objects they make up (e.g., ``mirror,'' which is actually an object, is identified as a material by finding mirrors, ``leather'' is recognized by finding sofas and ottomans, and ``fabric'' is identified by recognizing pillows).

On the other hand, there have been recent attempts to study the separation of materials from other image context. Hu~\etal\cite{Hu2011} briefly investigated the correlations between objects and materials. Schwartz and Nishino~\cite{Schwartz2013,Schwartz2015} proposed to recognize materials from small local image patches taken from inside the boundary of object regions and do not contain any shape cues to achieve per-pixel recognition based on visual material traits (e.g., ``smooth,'' ``hard,'' and ``organic'') as discriminative internal representations that transcend material categories. Although these methods successfully demonstrate the power of local material recognition, they do not show the integration of rich image context for combined reasoning of materials in images.

Global image context, including both what the object is and where it is, can provide rich information to narrow down what things and stuff are made of. A city street, for example, is likely to contain asphalt, rubber, glass, and metal materials. Figure~\ref{fig:conditional-distributions}
shows examples of actual correlations between materials and context given ground-truth objects from the MS COCO database~\cite{Lin2014}, along with global context in the form of place category predictions from the MIT Places CNN~\cite{Zhou2014}. The challenge in exploiting these contextual cues is that if we were to simply attempt to obtain exhaustive annotations for materials, objects and places, we would be searching a product space with an extremely large number of combinations.

In this paper, we introduce a material recognition framework that fully leverages local appearance and global contextual cues to recognize materials at each pixel in an image. By relying on accurate global context that may be extracted from any image, we avoid having to obtain annotations for $\left\{ Places\right\} \times\left\{ Objects\right\} \times\left\{ Materials\right\} $. Materials are an inherently local property, and as such we aim to produce dense per-pixel material category predictions. We separate materials from objects and other context by training a full-resolution dense-output material CNN on small local image patches. We then introduce explicit accurate context cues, in the form of object and place category predictions, to higher levels of the network. This allows us to provide accurate context rather than the implicit and uncertain context present in large patches, and also avoids the tradeoff between spatial resolution and context observed in~\cite{Bell2015}. 

We group the context categories into a logical hierarchy and investigate the effect of hierarchical context level on material recognition. We find that each additional form of context we introduce provides an independent increase in material recognition accuracy, and that finer-grained context is better for material recognition. We also investigate the ideal level, in terms of the hierarchical levels of the CNN recognition framework, at which to introduce context. Intuitively, objects and places are high-level concepts. Our results agree with this, as we find that context is best used when it we introduce it at the highest level of the network.

Our results show that the explicit separation and (re-)integration of image context significantly improves local material recognition accuracy. We quantitatively evaluate the accuracy of our method and find that it outperforms previous approaches that implicitly mix materials with object and place context. On a recently introduced comprehensive and diverse material database \cite{Schwartz2016}, we confirm that our method achieves state-of-the-art accuracy with significantly less training data compared to past methods.


\section{Related Work}

Material recognition is usually done at an image patch level. These
patches are, however, relatively large in most existing works: they
span a significant area of the scene, sometimes even the entire image,
covering parts of or entire objects. Sharan~\etal\cite{Sharan2009}
introduced the earliest form of such classification with the Flickr
Materials Database (FMD). In the FMD, the image patch is the entire
image, and each image contains a single primary material of interest,
similar to image classification. Recently, Bell~\etal\cite{Bell2015}
demonstrated per-pixel material classification
using a large-scale annotated training data, the Materials in Context (MINC) dataset, and
a combination of CNN and CRF models for classification. Their method
uses a large image patch for each pixel, roughly a quarter of the entire
image, which inevitably mix in object or place
context to material appearance. This naturally leads to the reliance on recognizing the object to recognize the material, which would fundamentally necessitate extremely large training data that span the product space of materials and objects (and places). It is also important to point out that the dataset is highly biased as it is predominantly sources from professional real-estate photographs. Wang~\etal\cite{Wang2016} also demonstrate accurate dense
per-pixel material predictions using 4D light field images. Zhang~\etal\cite{Okatani2015}
have recently shown impressive performance on the FMD, but their results focus only on single patch predictions. These methods
mix materials and context interchangeably throughout the recognition pipeline, when they would be better-used in a factorized form (as we show).

Dense prediction, outputting a value or category prediction for each
pixel, has been extensively studied in the context of object recognition
and object semantic segmentation. Object recognition datasets, such
as ImageNet~\cite{Russakovski2015} or MS COCO~\cite{Lin2014},
often contain many (80-10,000) categories. Despite this, state-of-the-art
semantic segmentation methods such as DeepLab~\cite{Chen2016} focus
on only a small subset of coarse-grained categories. A notable and
relevant exception is the recent ADE20k dataset, scene parsing challenge,
and associated models~\cite{Zhou2016}. The dataset contains many
fully-segmented images, and the challenge defines a set of 150 categories
for semantic segmentation. We are not merely performing 
semantic segmentation. We instead aim to produce dense material 
predictions. For this, we find the ADE20k models to be ideal sources 
of per-pixel object category context information.

The use of context as a means to reduce ambiguity, whether in materials
or other cases, appears promising. Hu~\etal\cite{Hu2011} showed
that a simple addition of object category predictions as features
could potentially improve material recognition. On an unrelated topic,
Iizuka~\etal\cite{Iizuka2016} use scene place category predictions
to improve the accuracy of greyscale image colorization. Our work,
in contrast to these previous methods, takes advantage of multiple
sources of context and investigates the ideal granularity of context
categories. Within the framework of a Convolutional Neural Network
(CNN), we evaluate how the hierarchical level at which we introduce
context influences the accuracy of the corresponding material predictions.

\section{Local Material Recognition}

We aim to leverage scene context, such as objects and places, to improve
dense per-pixel material recognition. Our first step is to ensure
that, when recognizing materials, we are in fact dealing with just
materials and not an implicit fusion of materials and context. Schwartz
and Nishino~\cite{Schwartz2013,Schwartz2015} have proposed to achieve
such a separation by recognizing materials using only small local
image patches inside the boundary of objects as input, thereby avoiding
any influence from context derived from object shape or other global
features. Equally important, they have recently introduced a dataset aimed at local
material recognition, with carefully-selected categories chosen from
a material hierarchy~\cite{Schwartz2016}, and material annotations
that respect object boundaries. 
The taxonomy of
materials is based on canonical categorization defined in materials science~\cite{matbase} and material regions are carefully segmented for images sourced from a variety of databases including PASCAL
VOC database~\cite{Everingham2010}, the Microsoft COCO
database~\cite{Lin2014}, the FMD~\cite{Sharan2009}, and the ImageNet
database~\cite{Russakovski2015}. Although the total number of images (about 3000) are smaller compared to past datasets \cite{Bell2015}, the clean separation of materials and other context (i.e., objects and places), the principled material category definitions that avoid mixing objects and materials, and the additional care taken to minimize bias in types of images make it ideal for studying local material recognition. We are able to extract more than 200,000 image patches without object context (e.g., object boundaries) lurking in, which makes it a sufficiently large-scale dataset for training a local material recognition classifier.

Our goal is to integrate materials with context that may be partially
global (objects with large spatial extent but defined boundaries)
or fully global (scene place categories, one per image). The frameworks
introduced in~\cite{Schwartz2013,Schwartz2015} are only able to
make dense pixel-wise predictions in a sliding window fashion, and
do not offer any logical point at which to introduce global context.
To address this, we build a fully-convolutional CNN architecture,
based on the VGG-16 network of Simonyan and Zisserman~\cite{Simonyan2015}
with modifications to enable us to output dense full-resolution material
predictions with integrated global context. Bell~\etal\cite{Bell2015}
have previously investigated a similar architecture for material recognition.
They, however, rely on large (24\% image size) training patches and
a loosely-defined set of material categories (e.g., carpet as a material)
that collectively mix up materials and objects. In contrast, all of
our training is done with small local material image patches. Section~\ref{subsec:Model} describes the model architecture.

\section{Distributions of Materials and Context}

We have an intuitive understanding that, if one knows an object is,
for example, made of metal, then it may be a knife or a car, but probably
not a piece of clothing. Likewise, if we know an object is a cup,
then it is likely made of glass, plastic, or ceramic. We can quantitatively
evaluate the informative nature of context, such as object and place
categories, by computing the conditional probability distributions
of materials given each possible category of context. If our intuition is correct, then these distributions should be discriminative (e.g., have a low entropy relative to the corresponding discrete uniform distribution).

\subsection{Object Context}

\begin{figure}
\begin{centering}
\includegraphics[width=0.49\columnwidth]{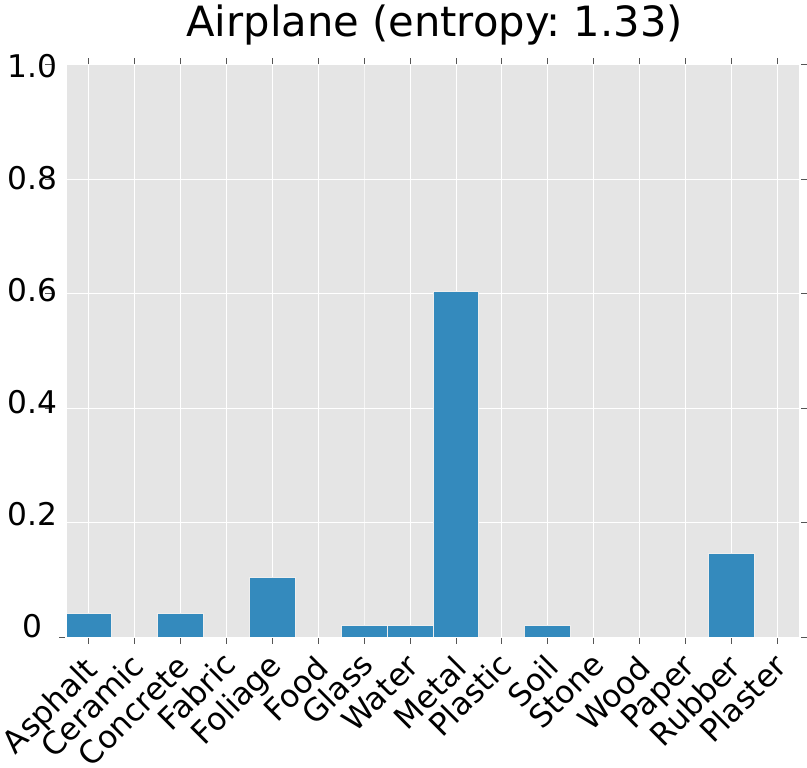}\includegraphics[width=0.49\columnwidth]{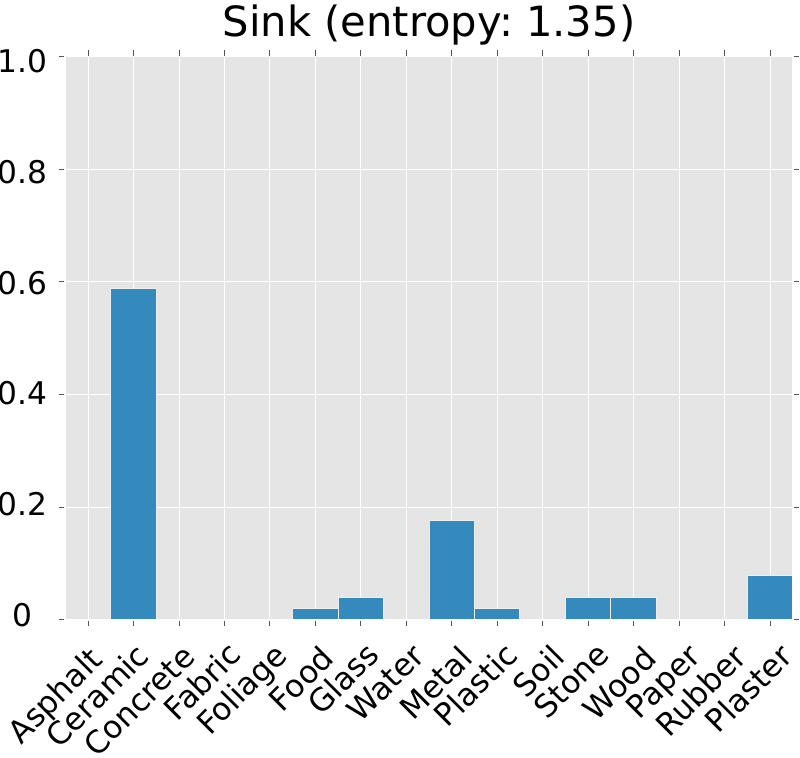}
\par\end{centering}
\vspace{5pt}

\begin{centering}
\includegraphics[width=0.49\columnwidth]{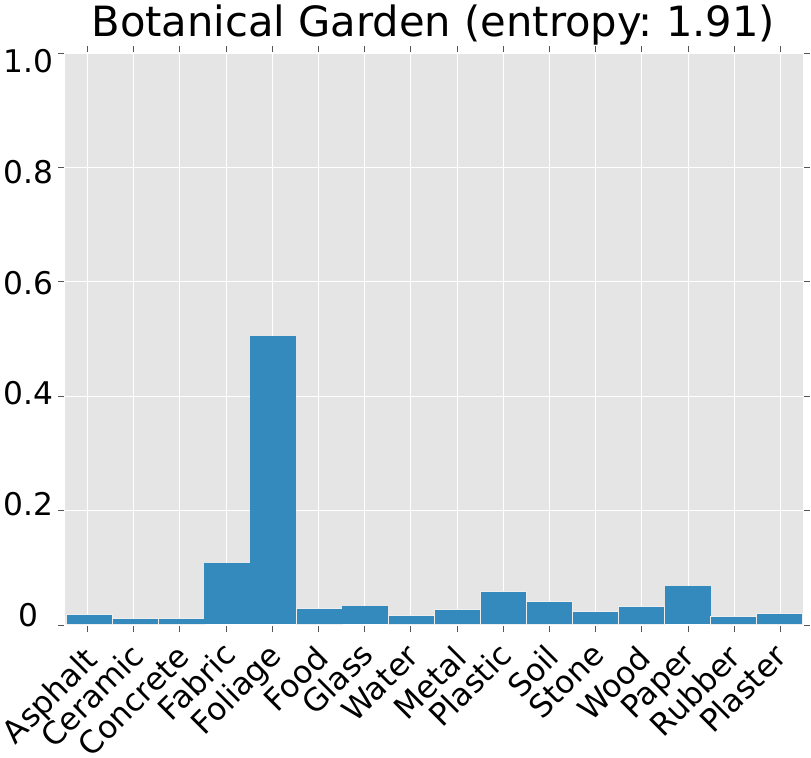}\includegraphics[width=0.49\columnwidth]{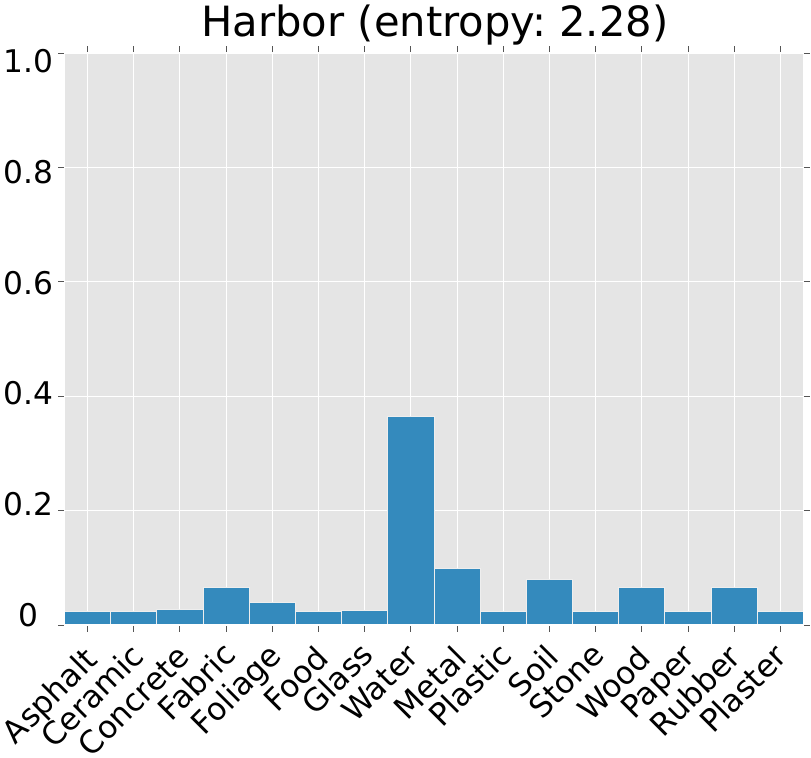}
\par\end{centering}
\caption{\label{fig:conditional-distributions}The conditional distributions
of materials given ground-truth object categories (top row) and predicted
places (bottom row) are highly discriminative. Many context categories
exhibit only a small set of materials. Some outliers are inevitable
as the ground-truth COCO segmentation masks do not perfectly conform
to actual object boundaries in the image. Places do not offer the
same very strong divisions of materials as objects do. Their distributions
are, however, still valuable as shown both by their entropy and the
resulting material recognition accuracy based on place context.}
\end{figure}

We can get an initial idea as to how discriminative context is by
using ground-truth object masks and corresponding materials to compute
the conditional distribution $p\left(M|O\right)$, where $M$ is the material category and $O$ is the object category. We use the material
database of~\cite{Schwartz2016} as they include images from databases
that contain object category map annotations (particularly, MS COCO~\cite{Lin2014}).
To compute the conditional probabilities, we take each image with
material annotations and find the object exhibiting each material
as indicated by the COCO ground truth. Figure~\ref{fig:conditional-distributions}
shows conditional material probabilities $p\left(M|O=o\right)$ for
a few selected object categories $o$. The entropy for the discrete
uniform distribution over 16 categories is 2.77, and as shown in Figure~\ref{fig:conditional-distributions} the
entropy given true object categories is much lower.

\subsection{\label{subsec:Place-Context}Place Context}

Objects are defined somewhat locally (at the level of groups of pixels, but still globally compared to the local material appearance we model)
and tend to exhibit only a small set of materials. In contrast, places
are single scene-wide properties and can encompass many objects and
materials. Despite this, we expect that places can still provide useful
cues to disambiguate local materials. Ceramic and paper, for example,
are often both flat white surfaces. Without seeing a specular highlight,
it may be difficult to distinguish the two given only a small local
patch. If, however, we know that the image patch originates from an
image of a classroom, it is more likely that the patch contains paper.

We can evaluate the discriminative power of places by using predictions from the MIT Places CNN~\cite{Zhou2014}. Figure~\ref{fig:conditional-distributions}
contains examples of the conditional distributions $p\left(M|P=p\right)$
for a few choices of place category $p$. While they are not uniformly
as discriminative as object categories, they still do provide some
useful cues. Botanical gardens, for example, tend to contain plants
as one would expect, and images of crosswalks contain asphalt, metal,
and rubber (roads, cars).

At least in the case of objects, and perhaps places as well, the conditional
distributions are so discriminative as to suggest that we might simply
multiply these distributions with the predictions of an existing material
recognition model and achieve improved accuracy. We initially investigated
this applied to a material recognition CNN as a baseline for comparisons. We, 
however, found that simple multiplication made a negligible difference
in the accuracy. This is due to the fact that many of the mispredictions of materials 
we might hope to correct are too strongly-predicted for simple multiplication
to have any effect.

\section{Local Materials and Global Context}

While the context cues from objects and places are highly discriminative,
we cannot simply treat them as a prior on material occurrence and
multiply them with a model's prediction. The model must instead have
the context available during training, so that the context may influence
the material predictions. Such an observation is consistent with the general
idea of leveraging top-down feedback with bottom-up recognition, for
instance, as demonstrated with object detection~\cite{Epshtein2008}
and human pose estimation Carreira~\etal\cite{IEF2015human}. Here,
we are obtaining the top-down information in the form of object and
place context. We treat the set of predicted context category probabilities,
obtained from state-of-the-art networks for scene recognition and
object semantic segmentation, as an additional feature in a dense
per-pixel material CNN. By concatenating these probabilities with
the high-level features in the network prior to output, we may take
full advantage of the strong material recognition cues available in
global image context.

\subsection{\label{subsec:Model}Context Integration Network}

\begin{figure*}
\begin{centering}
\includegraphics[width=2\columnwidth]{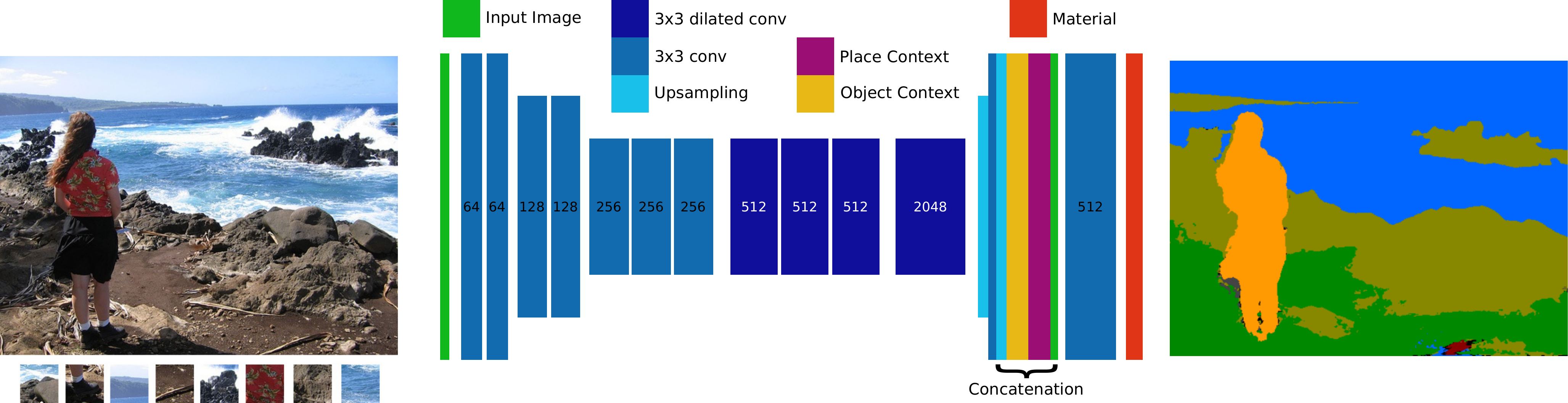}
\par\end{centering}
\caption{\label{fig:model-arch}We integrate local materials with scene context in
a CNN framework based on the VGG-16 architecture. We employ fewer
pooling steps to allow us to train on small local image patches, and
we add dilated convolution and upsampling (output-strided convolution a.k.a ``deconvolution'') layers to produce dense output. Per-pixel global context, in the form of place
and object category probabilities predicted from the input image via
existing models, is treated as an additional feature and concatenated
with the upsampled result prior to final prediction. For non-localized
context like place categories, we broadcast the probabilities across
the entire image. For improved spatial resolution, we also add a skip
connection between the input image and final output.}

\end{figure*}

As shown in Figure~\ref{fig:model-arch}, our model is based on the
VGG-16 architecture of Simonyan and Zisserman~\cite{Simonyan2015}
with a few fundamental modifications to enable dense prediction from local
material image patches with added global image context. To enable dense prediction,
we train a fully-convolutional form of the network where all fully-connected
layers are replaced with convolutions. We also add a series of upsampling
layers (output-strided convolution, also called ``deconvolution'') to match the input and output resolutions.

The level of downsampling in the original VGG-16 network
is not compatible with local material recognition. The minimum patch
size is constrained by the downsampling factor, and we would like
to use small image patches to maximize the
separation between local materials and global context. We find $48\times48$px patches represent an appropriate trade-off between eliminating all non-local information (single pixel ``patches'') and the large patches used by previous methods. While extremely small objects may still appear in such patches, relying on the database of~\cite{Schwartz2016} ensures that these objects are not labeled and will not create any undesired dependence on implicit context. To avoid too much downsampling, we remove the last set of pooling and filtering layers from the network. The remaining top two pooling layers are also removed and the corresponding layers above are replaced with dilated convolutions. In order to train on datasets where densely-segmented material ground truth may not always be available, we compute the softmax loss function at each pixel. The loss function is only evaluated at pixels where there is a known material. In this way we are able to take advantage of segmented material images without requiring completely dense annotation.

We leverage other contextual cues, namely object recognition and place recognition results, by running state-of-the-art classifiers separately and then integrating the results into the material recognition pipeline. This enables the complete separation of global image context recognition from local material recognition, which avoids requiring prohibitively large numbers of samples of the product space of materials, objects, and places as training data. It also allows us to use any source of object and place context without being limited to specific requirements for integration. 

Specifically, for global image context integration, we treat estimated context category probabilities
as additional features and concatenate them with existing features
in the network. For places, we only have 
a single set of category probabilities for the entire image. We replicate
these values across the image to match the image dimensions at the
level where the context is introduced. This is similar to the colorization work of Iizuka~\etal\cite{Iizuka2016}. We, however, introduce per-pixel context in the form of object semantic segmentation and find quantitative support for the spatial resolution and introduction level in the network of the added context.

\subsection{Extracting Per-Pixel Global Context}

We leverage global image context in the form of per-pixel object category predictions (semantic
segmentation) and single scene-wide place category predictions. For
places, we use the MIT Places CNN~\cite{Zhou2014}. A key feature
of their places database is that the categories are hierarchically
organized. As we will show below when we investigate the effects of
place category granularity, the number of context categories is important:
fine-grained context provides more useful information than simple,
higher-level context groups. In contrast, most existing object semantic segmentation
methods train their models on only a relatively small set of high-level
object categories. A notable exception is the ADE20k dataset~\cite{Zhou2016}.
The dataset contains over 2,000 object categories. The MIT Scene Parsing
Challenge, which relies on this dataset, selects 150 categories for semantic
segmentation. We use these categories and trained models for our per-pixel
object context.

\subsection{Hierarchical Place Context}

\begin{table}
\begin{centering}
\begin{tabular}{|c|c|c|}
\hline 
Hierarchy Level & Accuracy & Entropy\tabularnewline
\hline 
\hline 
High Level & 61.0\% & 2.51\tabularnewline
\hline 
Mid Level & 62.9\% & 2.40\tabularnewline
\hline 
Low Level  & 64.1\% & 2.27\tabularnewline
\hline 
All Places & 68.2\% & 1.91\tabularnewline
\hline 
\end{tabular}
\par\end{centering}
\vspace{5pt}

\caption{\label{fig:acc-vs-place-hierarchy}Place categories have an associated hierarchy
that are grouped by various attributes such as indoor/outdoor or manmade/natural. Fine-grained places may not appear in many images, but coarse grained categories may offer little in the way of material recognition cues. We in fact find that despite this, the finest category granularity offers the best material recognition performance. In this case, the 205 place categories are both fine-grained and sufficiently well-distributed across training examples.}
\end{table}

As part of their SUN database for scene and object recognition, Xiao~\etal\cite{Xiao2010}
define a hierarchy of place categories. This hierarchy raises the
question of whether any particular context granularity is more or
less useful for material recognition. On one hand, having an extremely
fine set of place categories might mean that few training examples
would appear from certain places. At the other extreme, the coarsest
division of places could only provide very general cues as to which
materials may be present.

To evaluate the importance of place granularity, we compute material
recognition accuracy scores using only place context at each level
of the SUN places hierarchy. We adapt their hierarchy to the place
categories recognized by the MIT places CNN and treat nodes within
each level of the hierarchy as place categories. The highest level
is the simple division of indoors vs.~outdoors, mid-level categories
deal with distinctions such as commercial and residential buildings,
or mountains and forests, and the lowest level includes smaller groups
such as entertainment or religious places. Results in Table~\ref{fig:acc-vs-place-hierarchy}
show that accuracy increases with place category granularity: more
detailed place categories provide more discriminative information
for material recognition. Computing the entropy of the conditional
distributions $p\left(M|\mathrm{P_{i}}\right)$ for place category
set $P_{i}$ at hierarchy level $i$ supports these results.

\subsection{\label{subsec:Context-Integration-Level}Integration Level and Spatial Resolution}

Existing methods for integration of context, such as that of~\cite{Iizuka2016}, find that adding context at the highest levels of the network results in a successful integration. This may make intuitive sense, as we want the context to directly inform existing category predictions at the end of the network. We show quantitatively that higher levels are indeed better suited for the addition of context. Additionally, as our method relies on object context that varies spatially (in contrast with single place context values across the entire image), we also investigate the effect of the object context's spatial resolution. As the context introduction level and spatial resolution are related (due to pooling layers), any observed change in accuracy could be caused by either the drop in spatial resolution or the change in level. We vary the effective resolution of the context by fixing the level and actual resolution (the number of pixels in the context map) then downsampling and upsampling the context.

Results in Table~\ref{tab:acc-vs-cnn-lvl} show that the highest level is 
indeed the ideal place at which to introduce global context. If introduced 
at lower levels, the network is free to overfit to the context and poor 
accuracy results. For this experiment we randomly initialized the weights of the network, thus the accuracy values are not comparable with results in later sections. If the network was initialized from pre-trained weights, as in our final results, then the accuracy would be artificially reduced as we introduced context at lower levels as the added context would invalidate the pre-trained weights above it. To separate the effects of spatial resolution and introduction level, we trained the same network with object context effective spatial resolution $\frac{1}{d},d\in\left\{2,4,8,16\right\}$, with context introduced after upsampling. Accuracy was essentially unaffected (71.4\% with 16$\times$ downsampling), showing that it is indeed the hierarchical network level and not the spatial resolution that determines the accuracy.

\begin{figure*}
\begin{centering}
\includegraphics[width=2\columnwidth]{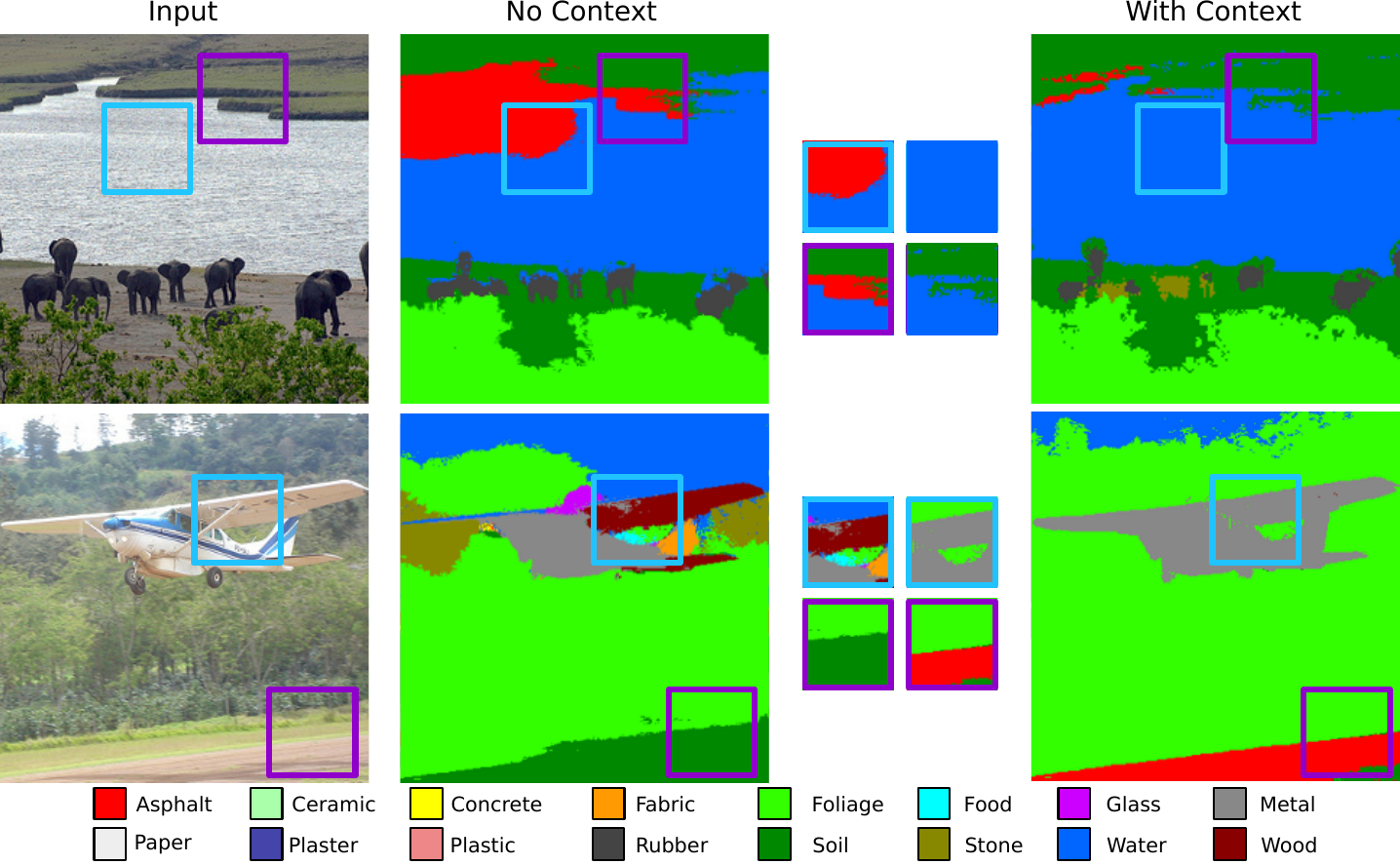}
\par\end{centering}
\caption{\label{fig:noctx-vs-ctx}These examples show that context helps disambiguate materials when local information is not sufficient. In the first set of insets, the water has a local appearance similar to asphalt. Global context suggests that this is unlikely. In the second set, we see that the airplane body is incorrectly recognized due to the lack of characteristic specular reflection that locally identifies metal. Again, context fixes this error. Sky is not a material and in this case has the local appearance of water, hence the prediction for those pixels in the second row.}
\end{figure*}

\begin{table}
\begin{centering}
\begin{tabular}{|>{\centering}m{70pt}|>{\centering}m{70pt}|}
\hline 
Layer & Accuracy\tabularnewline
\hline 
\hline
pool1 & 57.2\%\tabularnewline
\hline
pool2 & 58.1\%\tabularnewline
\hline
conv3\_3 & 60.2\%\tabularnewline
\hline
conv4\_3 & 60.7\%\tabularnewline
\hline
upsampling & 62.1\%\tabularnewline
\hline 
\end{tabular}
\par\end{centering}
\begin{centering}
\vspace{5pt}
\par\end{centering}
\caption{\label{tab:acc-vs-cnn-lvl}We show that objects and places can offer highly discriminative cues for material recognition. The integration,
however, introduces a potential for overfitting by relying too heavily
on the context. By plotting material recognition accuracy at various
context introduction levels, we see that the best level for context
introduction is the penultimate layer in the network.}
\end{table}


\begin{table}
\begin{centering}
\begin{tabular}{|c|c|>{\centering}m{60pt}|}
\hline 
Context & Accuracy & Mean Class Accuracy\tabularnewline
\hline 
\hline 
None & 63.4\% & 60.2\%\tabularnewline
\hline 
Only Places & 68.2\% & 67.7\%\tabularnewline
\hline 
Only Objects & 67.0\% & 63.5\%\tabularnewline
\hline 
Places + Objects & \textbf{73.0\%} & \textbf{72.5\%}\tabularnewline
\hline 
\end{tabular}
\par\end{centering}
\begin{centering}
\vspace{5pt}
\par\end{centering}
\caption{\label{tab:acc-contributions}By comparing per-pixel average accuracy
with no context, as well as with each separate form of additional
context, we see that the two forms of context (objects and places)
each have a strong effect on the recognition of materials. The combination
of objects and places is also significantly higher than either of
the two alone, suggesting that knowing both objects and places provides
unique cues not present in either individual category group.}
\end{table}

Table~\ref{tab:acc-contributions} contains a breakdown of the contributions
for each form of global context. Most importantly, the contributions from
objects and places are similar and the combination of the two outperforms
either individual context source. This suggests that the objects and
places are providing unique sources of information and are both
critical to the accurate recognition of materials. A cup, for example,
may be made of glass or plastic. If, however, you are in a bar, then
it is much more likely to be made of glass.

\section{Dense Material Recognition}

\begin{figure*}
\begin{centering}
\includegraphics[width=2\columnwidth]{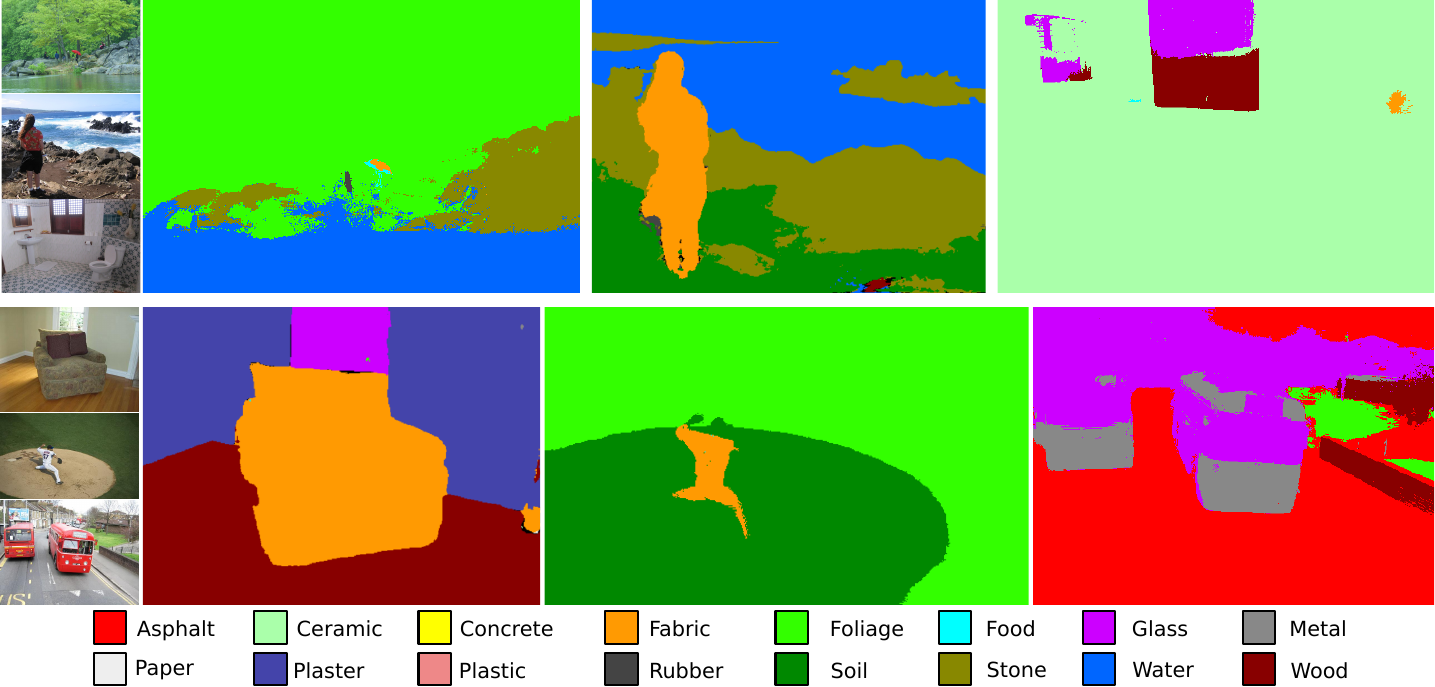}
\par\end{centering}
\caption{\label{fig:more-dense-preds}Additional examples of dense material recognition with global context show that the locally-recognized materials can also help disambiguate context. In the bathroom, all of the window has ``window'' as a context category, but the local appearance separates the glass from the wood. It is important to note that neither skin nor sky appear as materials in the definition of~\cite{Schwartz2016}. Skin is a unique case of a material that is visible only on one object category (people) in most databases, and the sky is not a material.}
\end{figure*}

\begin{table}
\begin{centering}
\begin{tabular}{|>{\centering}m{120pt}|c|>{\centering}m{60pt}|}
\hline 
Method & Accuracy & Mean Class Accuracy\tabularnewline
\hline 
\hline
VGG-16 + Upsampling & 66.7\% & 55.1\%\tabularnewline
\hline
MINC~\cite{Bell2015}, no retraining & 60.4\% & 67.5\%\tabularnewline
\hline
MINC~\cite{Bell2015}, retrained & 72.8\% & 70.1\%\tabularnewline
\hline
Ours (Places + Objects) & \textbf{73.0\%} & \textbf{72.5\%}\tabularnewline
\hline 
\end{tabular}
\par\end{centering}
\begin{centering}
\vspace{5pt}
\par\end{centering}
\caption{\label{tab:comparisons}Our initial argument was that large-patch-based methods can only implicitly use any available context and thus cannot make the best use of said context. We compare our per-pixel average material accuracy on the material database of~\cite{Schwartz2016} with the high-performing VGG16-based approach of~\cite{Bell2015} to show that explicit addition of context outperforms the implicit approach. As a baseline, we also train a version of our model based on the full VGG-16 model with 224x224 patches. As expected, this model is unable to take full advantage of the added context. As in~\cite{Bell2015}, we measure accuracy on overlapping categories only when not retraining.}
\end{table}

Results in Table~\ref{tab:comparisons} show that our approach outperforms the large-patch-based (VGG-16 + CRF) method of Bell~\etal\cite{Bell2015} on the local materials database of~\cite{Schwartz2016}, despite the fact that their model was trained on both millions of patches and fine-tuned on local material images. We evaluated all of the MINC-trained models (AlexNet, GoogleNet and VGG-16) and found VGG-16 to be the most accurate for this comparison. As a baseline, we train the full VGG-16 architecture (with large patches) on the local materials database. As expected, it is unable to take full advantage of the implicitly available context context. This can be viewed as the baseline case when the MINC model was only trained on the local materials database \cite{Schwartz2016}.

The MINC model's accuracy is also significantly lower on the local materials database \cite{Schwartz2016} compared to their own database, even after the final category layer was re-trained for the categories of~\cite{Schwartz2016}. These results suggest that the local materials database contains more diverse and challenging images. A large part of the MINC database comes from real-estate photographs and thus are inevitably biased in materials. The fact that MINC with no retraining exhibits lower mean-class accuracy even on a smaller number of categories further supports this.

We attempted to generate an approximate comparison on the MINC database by splitting the segmented MINC test images and training on one portion. Due, however, to the disproportionately small amount of data that could be extracted (only 7000 segments, roughly equivalent to 63,000 patches, whereas the MINC method uses 2.5 million patches~\cite{Bell2015}), a fair comparison of the methods on this database could not be achieved. For reference, our method still achieves 68.5\%, which is a smaller drop in accuracy than MINC when comparing cross-dataset performance of the two models. This further shows the bias inherent in the MINC database. Although the vast single click training data that MINC \cite{Bell2015} is able to leverage certainly is an advantage of not separating material appearance from surrounding context, these numbers and our rigorous comparative experimental results summarized in Table~\ref{tab:comparisons} clearly show that our framework outperforms the MINC model \cite{Bell2015} with significantly less training data on a much more comprehensive and diverse dataset. Please see our supplemental material for the full implementation and training details of our final model.

We can readily see in Figure~\ref{fig:noctx-vs-ctx} that the context helps disambiguate materials that may be difficult to recognize from only local information. When metal does not exhibit specular highlights or reflections, as is the case with the airplane body, the flat white surface offers little in the way of local recognition cues. Knowing either that the scene is an airport or that the current pixel belongs to a plane removes this ambiguity. Likewise, in the natural scene with elephants, the combination of high-frequency waves and specular reflection causes the water to appear like concrete. Scene context makes it clear that concrete would be unlikely in this case. In general, the predictions are accurate subject to the limitations of the training data. Skin is not a material in the dataset of~\cite{Schwartz2016}, and thus skin is often classified as the surrounding fabric. Sky is not a material and the predictions for sky are determined largely by context (ex.~metal at airports, water over the ocean). Additional qualitative examples in Figure~\ref{fig:more-dense-preds} show that the combination of local materials and global context results in accurate material predictions in the face of local ambiguity (both in the context and in the local appearance). Our supplemental material contains further examples of dense material predictions from our framework.

\section{Conclusion}

Our results show that we can successfully separate materials from their surrounding context and combine those materials with highly-discriminative forms of global context. Such a combination outperforms previous methods which implicitly rely on context being available in a large input image patch. Additionally, we performed a detailed investigation into the ideal granularity for context in material recognition as well as the hierarchical level and spatial resolution at which the  context should be introduced into a CNN framework.

The experimental results conclusively demonstrate that material recognition based on the explicit integration of local appearance and global context achieves state-of-the-art accuracy on a comprehensive and diverse dataset with less training data. We believe these results also suggest similar approaches to bottom-up top-down integration for other recognition tasks, which we are interested in exploring in future work.

\section*{Acknowledgements}
This work was supported by the Office of Naval Research grant N00014-16-1-2158
(N00014-14-1-0316)  and N00014-17-1-2406, and the National Science Foundation
award IIS-1421094. The Titan X used for part of this research was donated by
the NVIDIA Corporation.

\bibliographystyle{ieee}
\bibliography{iccv2017}

\end{document}